\documentclass[twoside,11pt]{article}

\usepackage[tbtags]{amsmath}
\usepackage{url}
\usepackage{graphicx}
\usepackage{subcaption}

\usepackage{jmlr2e}
\usepackage[version=4]{mhchem}

\usepackage{algorithm}
\usepackage{algpseudocode}

\usepackage{tikz}
\usepackage{tikzscale}
\usepackage{pgfplots}
\pgfplotsset{compat=1.9}

\DeclareMathOperator*{\argmin}{arg\,min}

\newcommand{\norm}[1]{\left\lVert#1\right\rVert}

\ShortHeadings{Prototypal Analysis and Prototypal Regression}{Wu and Tabak}
\firstpageno{1}

\begin{document}

\title{Prototypal Analysis and Prototypal Regression}
\author{\name Chenyue Wu \email chenyue@cims.nyu.edu \\
	\name Esteban G. Tabak \email tabak@cims.nyu.edu \\
	\addr Courant Institute of Mathematical Sciences\\
	New York University\\
	New York, NY 10012, USA}

\editor{}

\maketitle

\begin{abstract}
	Prototypal analysis is introduced to overcome two shortcomings of archetypal analysis: its sensitivity to outliers and its non-locality, which reduces its applicability as a learning tool. Same as archetypal analysis, prototypal analysis finds prototypes through convex combination of the data points and approximates the data through convex combination of the archetypes, but it adds a penalty for using prototypes distant from the data points for their reconstruction. Prototypal analysis can be extended---via kernel embedding---to probability distributions, since the convexity of the prototypes makes them interpretable as mixtures. Finally, prototypal regression is developed, a robust supervised procedure which allows the use of distributions as either features or labels.
\end{abstract}


\section{Introduction}
\label{sec:introduction}

Archetypal analysis, an unsupervised learning method introduced by
\citet{cutler1994archetypal}, approximates a  set of data points by convex combinations of archetypes, which are themselves convex combinations of the original data. At the cost of introducing convexity constraints into the optimization, archetypal analysis achieves interpretability, as a convex combination can be thought of as a weighted sum of its components---not so a general linear combination, where components can be subtracted as well as added. 
This extra computational cost can be handled efficiently, as several studies have shown \citep{bauckhage2009making,morup2012archetypal,chen2014fast}.

Archetypal analysis has been applied in physics \citep{stone1996archetypal,stone2002exploring,chan2003archetypal}, biology \citep{huggins2007toward,romer2012early,thogersen2013archetypal}, psychology \citep{thurau2011introducing,drachen2012guns,drachen2016guns,sifa2013archetypical}, marketing \citep{li2003archetypal,d2006archetypal}, performance analysis \citep{porzio2006archetypal,porzio2008use,eugster2012performance,seiler2013archetypal} and computer vision \citep{marinetti2006matrix,thurau2009archetypal,cheema2011action,asbach2013understanding,xiong2013face}.

Despite the many positive features of archetypal analysis, one can point out two significant drawbacks. One is its sensitivity to outliers: since the data is approximated by its projection on the convex hull of the archetypes, adding a point outside of the boundary of the data impacts the archetypes to a large degree. Another drawback of the methodology is its non-locality: data points are approximated as convex combinations of archetypes that may be very far away. For many learning tools, such as regression, such representation is of little use.


This paper introduces prototypal analysis as a robust alternative to archetypal analysis without these drawbacks.
Prototypal analysis preserves interpretability, as it finds prototypes via convex combinations of the data and reconstructs the data as convex combinations of the prototypes. The difference between archetypal and  prototypal analysis is that the former allows arbitrary convex combination of archetypes for representing the data, while the later penalizes the use of prototypes far away from a data point to represent it. Technically, this is achieved by adding a $L_1$ penalty term on the reconstructing coefficients for each point, with weights that depend on the distance between the point and the prototype under consideration. As a consequence, a point far away from the majority of the data would contribute little to the reconstruction and will not be chosen as a prototype.

The locality of the reconstruction by prototypes makes them useful for key learning tasks such as regression. Given training data on predictors and responses, regression concerns inferring the response for new instances of the predictors. We introduce prototypal regression as a new regression method with the advantage of interpretability and robustness. Prototypal regression uses convex combinations to extract prototypes from both the predictors and the response. The regression relationship is built with pairs of one prototype from the predictor and one prototype from the response, i.e. prototypal regression maps each prototype from the predictor to one prototype from the response and extends to all values of the predictors via local convex combinations. Here convexity is the source of interpretability and, combined with locality, of robustness, as an outlier will only affect the predictions in its immediate neighborhood.

Kernel methods and reproducing kernel Hilbert space (RKHS) are widely used in machine learning to extend algorithms where only inner products among data points are required \citep{scholkopf2002learning,shawe2004kernel,hofmann2008kernel}. This is the case of archetypal analysis, which can therefore be extended via kernels \citep{morup2012archetypal}. Examples of application can be found in time series clustering \citep{bauckhage2014kernel}, behavior analysis \citep{sifa2014playtime} and image processing \citep{zhao2015multiple,zhao2016bilateral}. Prototypal analysis and prototypal regression can be kernelized as well, enabling in particular the use of probability distributions as either features or outputs, in lieu of the more conventional discrete or real-valued scalars and vectors.  This extension is particularly well suited for archetypal and prototypal analysis, as their underlying convex combinations correspond to mixtures of distributions.
We adopt kernel embedding (also known as kernel mean embedding) to extend archetypal analysis, prototypal analysis and prototypal regression to handle distributional data. Kernel embedding maps probability distributions or their samples into a RKHS. Using the inner products of the RKHS, one can find archetypes and prototypes of distributions and also perform regression in this infinite dimensional setting. More generally, kernel embedding enables prototypal regression to deal with a blend of categorical, numerical and distributional data.

In prior work, \citet{muandet2012learning} extends support vector machine to support measure machine for classification of distributions using the kernel embedding induced inner product. \citet{szabo2015two,szabo2016learning} performs a similar extension for kernel ridge regression. \citet{poczos2013distribution} regresses numbers from distributions through a kernel-kernel estimator, which involves one kernel for density estimation and  another for kernel smoothing, using the distance between the distributions to weight the response variables. 
\citet{oliva2013distribution} introduces a distribution to distribution regression model via orthogonal series density estimation on the response distributions and kernel density estimation on the predictor distributions and the new input.

The rest of this paper is organized as follows: Section \ref{sec:archetypal analysis} briefly reviews archetypal analysis and empirically shows that it is not robust to outliers and that, as it concentrates on the boundary of the data, it does not resolve the underlying space well. Section \ref{sec:prototypal analysis} introduces prototypal analysis as a robust unsupervised method to find prototypes and build data-driven barycentric coordinates system without these two drawbacks. Section \ref{sec:prototypal regression} introduces simple and multiple prototypal regression---the latter applicable to features of different nature that cannot naturally be regarded as components of a vector. Section \ref{sec:kernel} extends archetypal and prototypal analysis and prototypal regression via kernels and applies it to the analysis of distributional data.

\section{Archetypal Analysis}
\label{sec:archetypal analysis}

Archetypal analysis approximates data points by convex combination of ``archetypes'', which are themselves convex combinations of the data points \citep[see][]{cutler1994archetypal}.
Given a data set $\{\mathbf{x}_i\}_{i=1}^{n}$, one seeks archetypes of the form
\begin{equation}
	\label{eqn:aa:archetype}
	\mathbf{u}_{j} = \sum_{i=1}^n b_{ij} \mathbf{x}_i, \quad
	\sum_{i=1}^n b_{ij} =1, \quad b_{ij} \geq 0, \quad j \in [1,k]
\end{equation}
and approximates each data point through 
\begin{equation}
	\label{eqn:aa:reconstruct}
	\mathbf{x}_{i} \approx \sum_{j=1}^k a_{ji} \mathbf{u}_j, \quad
	\sum_{j=1}^k a_{ji} =1, \quad a_{ji} \geq 0, \quad i \in [1,n],
\end{equation}
by solving the following optimization problem:
\begin{equation}
	\label{eqn:aa}
	\min_{\substack{a_{ji} \geq 0, b_{lj} \geq 0\\ \sum_{j=1}^k a_{ji} =1\\ \sum_{l=1}^n b_{lj} =1}}\ \sum_{i=1}^{n}{\norm{\mathbf{x}_i - \sum_{j=1}^{k} a_{ji} \sum_{l=1}^{n} b_{lj} \mathbf{x}_l}^2}.
\end{equation}

\begin{algorithm}[t]
  \begin{algorithmic}[1]
    \Require Data $\{x_{i}\}_{i=1}^{n}$, $k$: number of archetypes.
    \Ensure Archetypes $\{u_{j}\}_{j=1}^{k}$ and approximation $\{\hat{x}_{i}\}_{i=1}^{n}$ to data  by their convex combination.
    \State $\displaystyle (a_{ji}) , (b_{lj}) \gets \argmin_{\substack{a_{ji} \geq 0, b_{lj} \geq 0\\a_{1i} + \cdots + a_{ki}=1\\b_{1j} + \cdots + b_{nj}=1}}\sum_{i=1}^{n}{\norm{\mathbf{x}_i - \sum_{j=1}^{k} a_{ji} \sum_{l=1}^{n} b_{lj} \mathbf{x}_l}^2}$
    \For{$j=1,\cdots,k$}
    \State $u_{j} \gets b_{1j} \mathbf{x}_1 + \cdots + b_{nj} \mathbf{x}_{n}$
    \EndFor
    \For{$i=1,\cdots,n$}
    \State $\hat{x}_{i} \gets a_{1i} \mathbf{u}_1 + \cdots + a_{ki} \mathbf{u}_{k}$
    \EndFor
    \State \Return $\{u_{j}\}_{j=1}^{k}$, $\{\hat{x}_{i}\}_{i=1}^{n}$
  \end{algorithmic}
  \caption{Archetypal Analysis}\label{algorithm:aa}
\end{algorithm}

\begin{figure}[tb]
	\begin{center}
		\includegraphics[width=\textwidth]{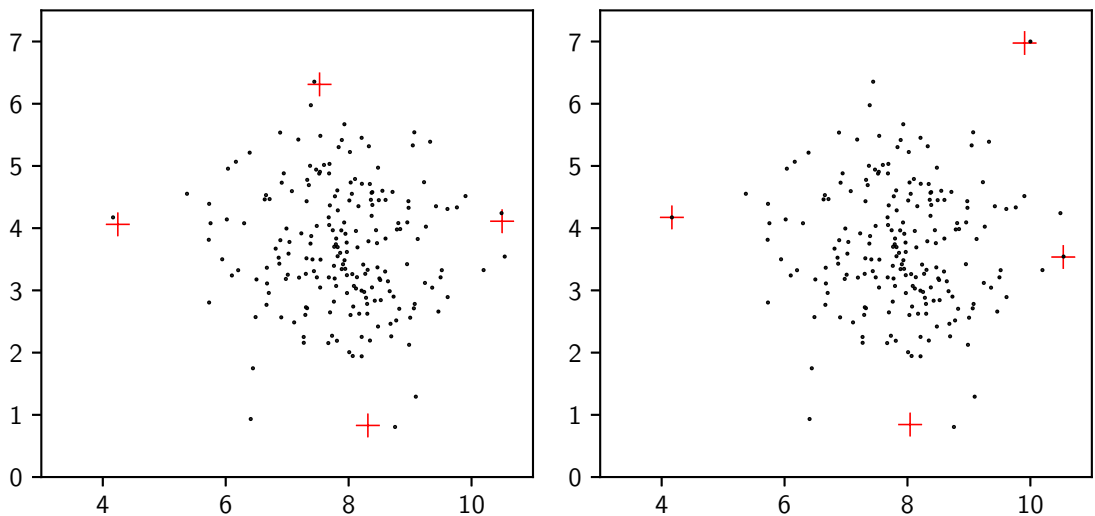}
	\end{center}%
	\caption[Archetypal analysis with outlier.]{Archetypal analysis on two dimensional data with 4 archetypes. Data of the right figure contains one more outlier than the left figure. The archetypes are visualized using the '+' sign. Adding one outlier fundamentally changes the location of the archetypes. In addition, the reconstruction of many data-points in terms of the archetypes is not unique.}
	\label{fig:aa_outlier}
\end{figure}

\begin{figure}[tb]
    \begin{center}
        \includegraphics[width=\textwidth]{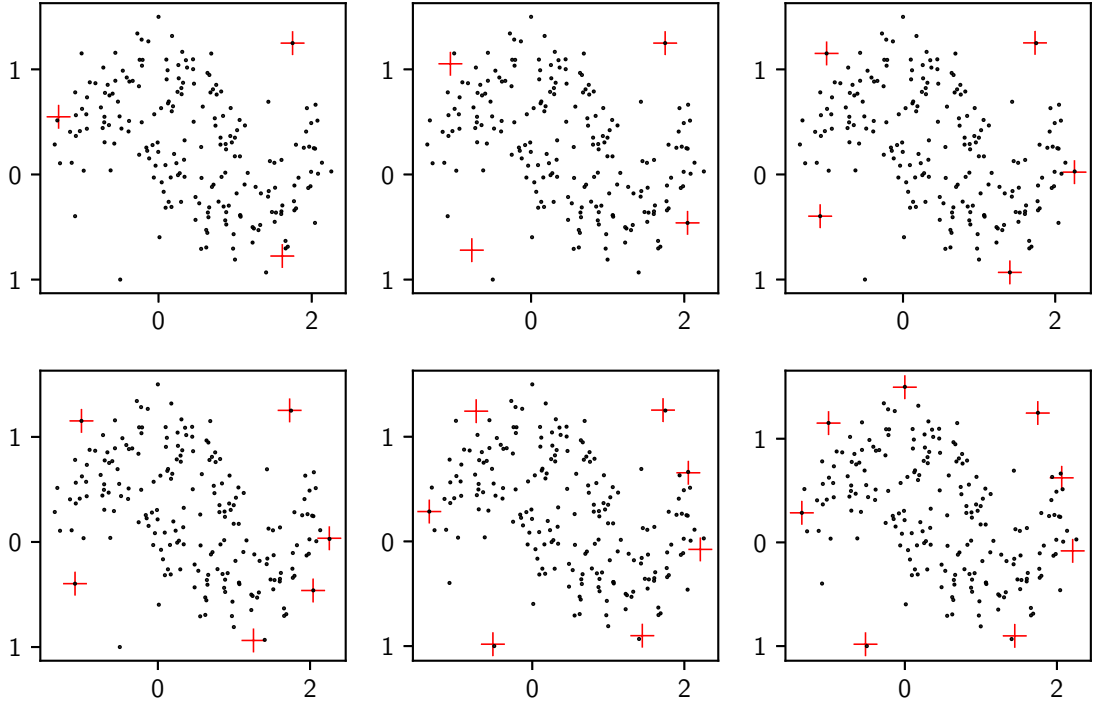}
    \end{center}%
    \caption[Archetypal analysis with different number of archetypes.]{Archetypal analysis on two dimensional data with 3, 4, 5, 6, 7 and 8 archetypes. The archetypes are visualized using the '+' sign. As the number of archetypes grows, they cover just the perimeter of the convex hull of the data.}
    \label{fig:num_archetypes}
\end{figure}

%
As archetypal analysis minimizes the distance between the data and the convex hull of the archetypes, it tends to choose as archetypes extreme points among the data in order to enlarge this convex hull. In particular, when the data includes outliers, these are typically chosen as archetypes, as illustrated in Figure \ref{fig:aa_outlier}. As the number $k$ of archetypes grows, they sit on the boundary of the convex hull of the data, not resolving its interior, as shown in Figure \ref{fig:num_archetypes}. Also, when $k$ is sufficiently large (typically when $k > d$, where $d$ is the number of the vertices of the convex hull spanned by $\{\mathbf{x}_i\}_{i=1}^{n}$), the $a_{ji}$ are not uniquely defined.

\section{Prototypal Analysis}
\label{sec:prototypal analysis}

\begin{figure}[tb]
	\begin{center}
		\includegraphics[width=\textwidth]{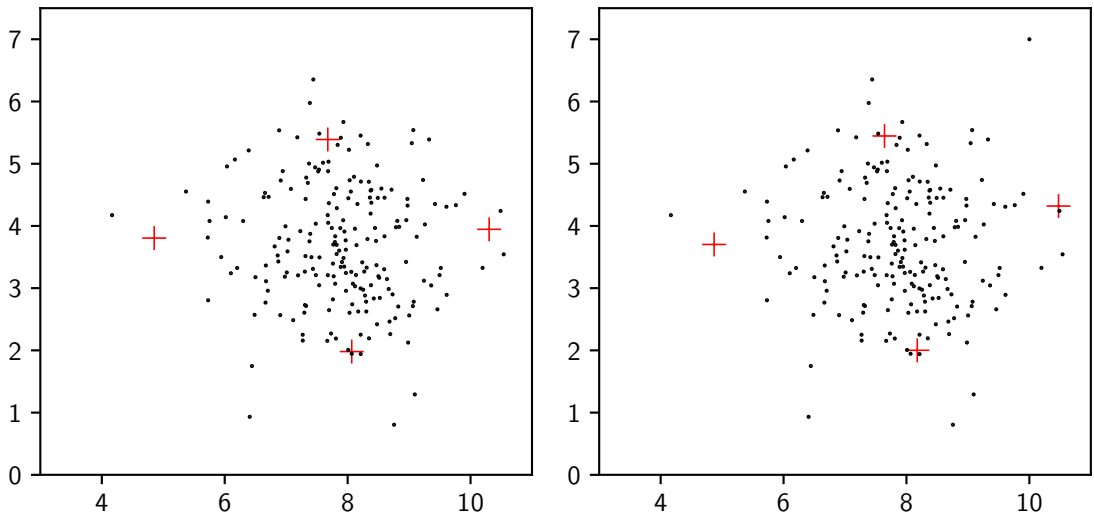}
	\end{center}%
	\caption[Prototypal analysis with outlier.]{Prototypal analysis on two dimensional data with 4 prototypes and penalty 0.05. The data of the right figure contains one more outlier than the left figure, but this affects the location of the prototypes only minimally. The prototypes are visualized using '+' signs.}
	\label{fig:pa_outlier}
\end{figure}

Like archetypal analysis, prototypal analysis finds prototypes $\{\mathbf{u}_{j}\}_{j=1}^{k}$as convex combinations of the data points $\{\mathbf{x}_i\}_{i=1}^{n}$, and approximates the latter as convex combinations of the former, as in Equation \ref{eqn:aa:archetype} and \ref{eqn:aa:reconstruct}. The difference lies in that, when reconstructing each data point, prototypal analysis is biased toward using prototypes near that point. To this end, it adds a penalty term on the distance between points and prototypes, replacing the objective function in Equation \ref{eqn:aa} by
\begin{equation}
	\label{eqn:pa}
	\min_{\substack{a_{ji} \geq 0, b_{lj} \geq 0\\ \sum_{j=1}^k a_{ji} =1\\ \sum_{l=1}^n b_{lj} =1}}\ \sum_{i=1}^{n}{\norm{\mathbf{x}_i - \sum_{j=1}^{k} a_{ji} \sum_{l=1}^{n} b_{lj} \mathbf{x}_l}^2} + \lambda \sum_{i=1}^{n}\sum_{j=1}^{k}a_{ji}\norm{\mathbf{x}_{i}-\sum_{l=1}^{n} b_{lj} \mathbf{x}_l}^2,
\end{equation}
where $\lambda \ge 0$ is a tuning parameter. In the penalty term, $a_{ji}$, the weight of the $j$-th archetype in the reconstruction of $\mathbf{x}_{i}$, is multiplied by $\norm{\mathbf{x}_{i}-\sum_{l=1}^{n} b_{lj} \mathbf{x}_l}^2$, the square of distance between data point $\mathbf{x}_{i}$ and the $j$-th prototype $\mathbf{u}_j = \sum_{l=1}^{n} b_{lj} \mathbf{x}_l$. Hence the closer $\mathbf{x}_{i}$ is to the $j$-th prototype, the more weight this prototype will be assigned in the reconstruction. Compared with archetypal analysis, which tends to use extreme points as archetypes, prototypal analysis has prototypes that resemble the original data. Hence it is less sensitive to outliers. Figure \ref{fig:pa_outlier} shows the prototypes corresponding to the same data of Figure \ref{fig:aa_outlier}. In this case, adding one outlier does not change the archetypes significantly.
In the computational procedure we use to minimize Equation \ref{eqn:pa}, we alternate between minimizing over the $a$ and $b$, which is also the procedure of choice in archetypal analysis \citep{cutler1994archetypal}.

\begin{algorithm}[t]
  \begin{algorithmic}[1]
    \Require Data $\{x_{i}\}_{i=1}^{n}$, number of prototypes $k$, penalty coefficient $\lambda$.
    \Ensure Prototypes $\{u_{j}\}_{j=1}^{k}$ and reconstruction of data by archetypes $\{\hat{x}_{i}\}_{i=1}^{n}$.
    \State $\displaystyle (a_{ji}) , (b_{lj}) \gets \argmin_{\substack{a_{ji} \geq 0, b_{lj} \geq 0\\a_{1i} + \cdots + a_{ki}=1\\b_{1j} + \cdots + b_{nj}=1}}\sum_{i=1}^{n}{\norm{\mathbf{x}_i - \sum_{j=1}^{k} a_{ji} \sum_{l=1}^{n} b_{lj} \mathbf{x}_l}^2} + \lambda \sum_{i=1}^{n}\sum_{j=1}^{k}a_{ji}\norm{\mathbf{x}_{i}-\sum_{l=1}^{n} b_{lj} \mathbf{x}_l}^2$
    \For{$j=1,\cdots,k$}
    \State $u_{j} \gets b_{1j} \mathbf{x}_1 + \cdots + b_{nj} \mathbf{x}_{n}$
    \EndFor
    \For{$i=1,\cdots,n$}
    \State $\hat{x}_{i} \gets a_{1i} \mathbf{u}_1 + \cdots + a_{ki} \mathbf{u}_{k}$
    \EndFor
    \State \Return $\{u_{j}\}_{j=1}^{k}$, $\{\hat{x}_{i}\}_{i=1}^{n}$
  \end{algorithmic}
  \caption{Prototypal Analysis}\label{algorithm:pa}
\end{algorithm}

Prototypal analysis can be viewed as a mixture of archetypal analysis and k-means clustering. When $\lambda$ goes to infinity, only the penalty term remains in prototypal analysis, and the problem reduces to
\begin{equation}
	\label{eqn:pa:penalty}
	\min_{\substack{a_{ji} \geq 0, b_{lj} \geq 0\\ \sum_{j=1}^k a_{ji} =1\\ \sum_{l=1}^n b_{lj} =1}}\ \sum_{i=1}^{n}\sum_{j=1}^{k}a_{ji}\norm{\mathbf{x}_{i}-\sum_{l=1}^{n} b_{lj} \mathbf{x}_l}^2,
\end{equation}
which is equivalent to K-means clustering, with the prototypes $u_j = \sum_{l=1}^{n} b_{lj} \mathbf{x}_l$ playing the role of barycenters. To see this equivalence, notice two facts about the solution to Equation \ref{eqn:pa:penalty}:
\begin{enumerate}
\item For each observation $\mathbf{x}_{i}$, the only nonzero $a_{ji}$ corresponds to the closest $u_j$, for which $a_{ji} = 1$.
\item For each prototype $u_j$, the only nonzero $b_{lj}$ correspond to those $l$ such that $u_j$ is the closest prototype to $\mathbf{x}_{i}$. Moreover, these $b_{lj}$ all have the same value, as the barycenter of a set of points is the minimizer of the sum of the square distances to them.
\end{enumerate}

\begin{figure}[tb]
    \begin{center}
        \includegraphics[width=\textwidth]{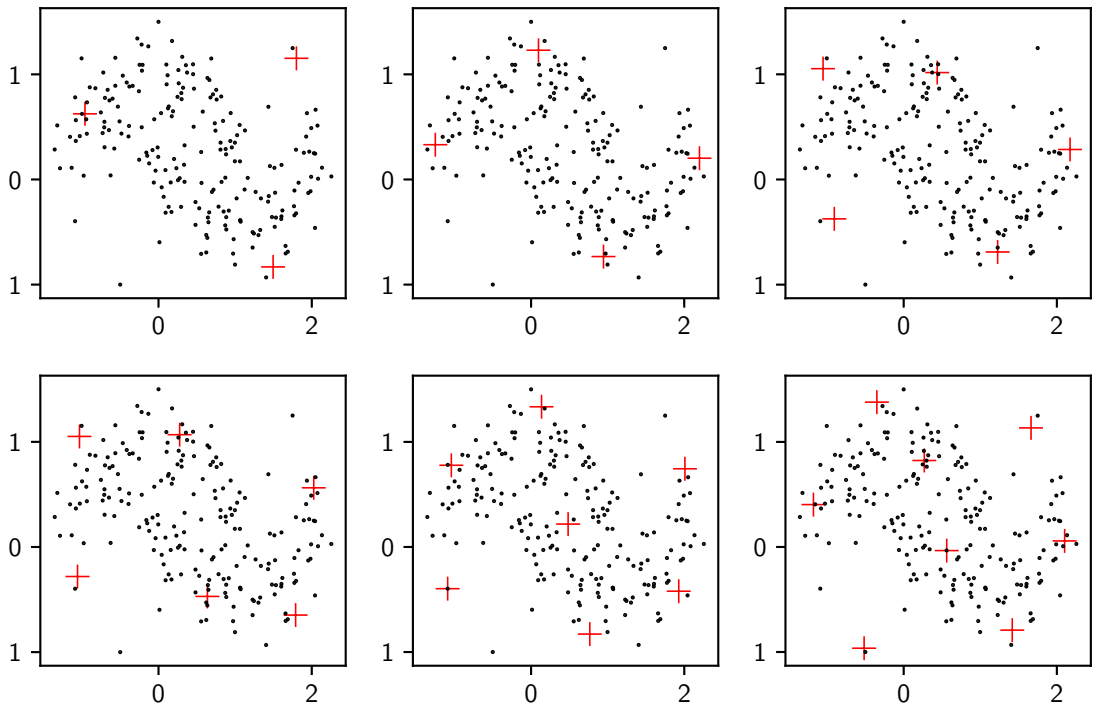}
    \end{center}%
    \caption[Prototypal analysis with different number of prototypes.]{Prototypal analysis on two dimensional data with penalty 0.05. The number $k$ of prototypes is set to 3, 4, 5, 6, 7 and 8. The prototypes are visualized using '+' signs. Unlike archetypes, as the number of prototypes grows, they populate all data-rich areas.}
    \label{fig:num_prototypes}
\end{figure}

\section{Prototypal Regression}
\label{sec:prototypal regression}

Given a set of predictor-response pairs $\left(\mathbf{x}_i,\mathbf{y}_i\right)$, regression is the task estimating the response $\mathbf{y}_0$ corresponding to a new value $\mathbf{x}_0$ of the predictor. Performing prototypal analysis on the $\left\{\mathbf{x}_i\right\}$ yields the prototypes $\left\{u_j\right\}$ and a rule that approximates $\mathbf{x}_0$ as a convex combination of a local subset of the $\left\{u_j\right\}$. Hence introducing prototypes $\left\{v_j\right\}$ in $\mathbf{y}$-space that approximate the images of the $\left\{u_j\right\}$, one can estimate $\mathbf{y}_0$ as the corresponding convex combination of the $\left\{v_j\right\}$.

\subsection{Simple Prototypal Regression}

\begin{figure}[ptb]
  \begin{center}
    \input{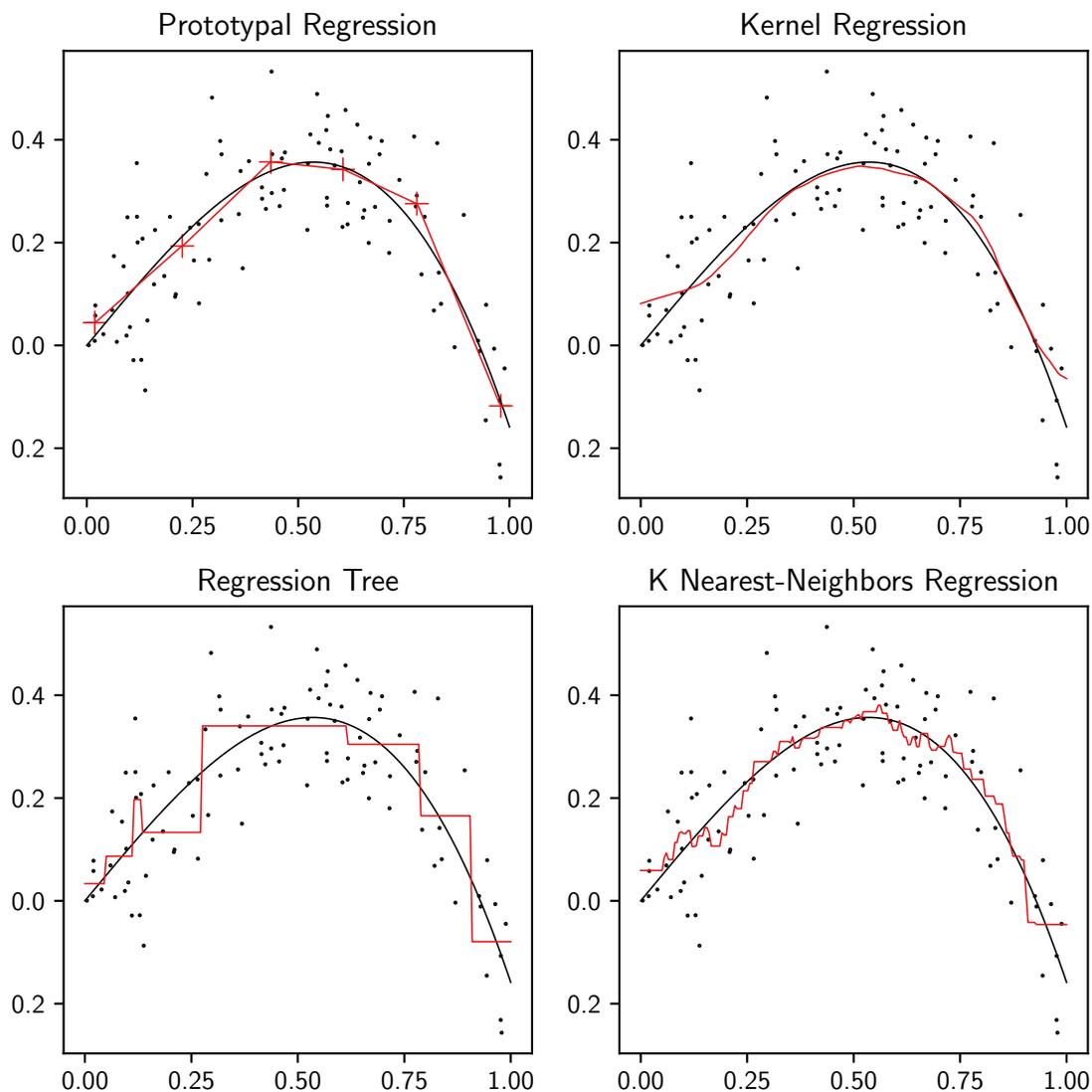}
  \end{center}
  \caption{100 pairs $x_i$, $y_i$ are sampled from a Gaussian conditional distribution with conditional mean $\bar{y} = \text{sin}(x) - x^3$ (the black curve), $x \sim U[0,1]$, $y = \bar{y} + \epsilon, \epsilon \sim \mathcal{N}(0,0.1)$. The red curves arise from regression. Top left panel: prototypal regression with 6 prototypes and penalty 0.01 (The prototypes of $x_i$ and $y_i$ are visualized using '+' signs.) Top right panel:  kernel regression with Epanechnikov kernel with (half) window width $\lambda=0.15$. Lower left panel: regression tree. Lower right panel: 10-nearest-neighbor regression.}
  \label{fig:toy_regression}
\end{figure}

Simple prototypal regression estimates the response $\mathbf{y}$ from a single predictor $\mathbf{x}$, where both predictor and response can be vectorial, using prototypes of both $\mathbf{x}$ and $\mathbf{y}$.
The prototypes of $\mathbf{x}$ come directly from prototypal analysis, i.e. solving Equation \ref{eqn:pa}, while the choice of prototypes of $\mathbf{y}$ takes the regression into account. Denoting by $\mathbf{u}_{j}$ the prototypes of $\mathbf{x}$ and by $\mathbf{v}_{j}$ the prototypes of $\mathbf{y}$, the prototype pair $(\mathbf{u}_j, \mathbf{v}_j)$ defines the regression function $\hat{\mathbf{f}}$ via
\begin{equation}
  \label{eqn:spr:predict}
  \hat{\mathbf{f}}(\mathbf{x}_0) = a_{10} \mathbf{v}_{1} + \cdots + a_{k0} \mathbf{v}_{k}
\end{equation}
where $a_{j0}$ are the barycentric coordinates of $\mathbf{x}_0$ in prototypal analysis:
\begin{equation}
  \label{eqn:spr:reconstruct}
  \min_{\substack{a_{j0} \geq 0\\\sum_{j=1}^k a_{j0}=1}}\norm{\mathbf{x}_0 - \sum_{j=1}^{k} a_{j0} \mathbf{u}_j}^2 + \lambda \sum_{j=1}^{k}a_{j0}\norm{\mathbf{x}_{0}- \mathbf{u}_{j}}^2.
\end{equation}
Given the weights $\left\{a_{ji}\right\}$ for reconstructing $\mathbf{x}_{i}$ in terms of the $\left\{\mathbf{u}_{j}\right\}$, the prototypes $\mathbf{v}_j$ are obtained by minimizing the squared errors of Equation \ref{eqn:spr:predict} on $(\mathbf{x}_i,\mathbf{y}_i)$, i.e. 
\begin{equation}
  \label{eqn:spr:prototype_y}
\mathbf{v}_j = \sum_{i=1}^{n} c_{ij} \mathbf{y}_i, \quad  c = \argmin_{\substack{c_{lj} \geq 0\\\sum_{i=1}^n c_{ij}=1}}\sum_{i=1}^{n}{\norm{\mathbf{y}_i - \sum_{j=1}^{k} a_{ji} \sum_{l=1}^{n} c_{lj} \mathbf{y}_l}^2}.
\end{equation}
Figure \ref{fig:toy_regression} illustrates simple prototypal regression, kernel regression, regression tree and k nearest-neighbor regression on a one-dimensional synthetic data set.

\begin{algorithm}[t]
  \begin{algorithmic}[1]
    \Require Predictor data $\{\mathbf{x}_{i}\}_{i=1}^{n}$, response data $\{\mathbf{y}_{i}\}_{i=1}^{n}$, number of prototypes $k$, penalty coefficient $\lambda$.
    \Ensure Prototypes $\{\mathbf{u}_{j}\}_{j=1}^{k}$ and $\{\mathbf{v}_{j}\}_{j=1}^{k}$ for predictor and response respectively.
    \State $\displaystyle (a_{ji}) , (b_{lj}) \gets \argmin_{\substack{a_{ji} \geq 0, b_{lj} \geq 0\\a_{1i} + \cdots + a_{ki}=1\\b_{1j} + \cdots + b_{nj}=1}}\sum_{i=1}^{n}{\norm{\mathbf{x}_i - \sum_{j=1}^{k} a_{ji} \sum_{l=1}^{n} b_{lj} \mathbf{x}_l}^2} + \lambda \sum_{i=1}^{n}\sum_{j=1}^{k}a_{ji}\norm{\mathbf{x}_{i}-\sum_{l=1}^{n} b_{lj} \mathbf{x}_l}^2$
    \For{$j=1,\cdots,k$}
    \State $\mathbf{u}_{j} \gets b_{1j} \mathbf{x}_1 + \cdots + b_{nj} \mathbf{x}_{n}$
    \EndFor
    \State $\displaystyle (c_{lj}) \gets \argmin_{\substack{c_{lj} \geq 0\\c_{1j} + \cdots + c_{nj}=1}}\sum_{i=1}^{n}{\norm{\mathbf{y}_i - \sum_{j=1}^{k} a_{ji} \sum_{l=1}^{n} c_{lj} \mathbf{y}_l}^2}$
    \For{$j=1,\cdots,k$}
    \State $\mathbf{v}_{j} \gets c_{1j} \mathbf{y}_1 + \cdots + c_{nj} \mathbf{y}_{n}$
    \EndFor
    \State \Return $\{\mathbf{u}_{j}\}_{j=1}^{k}$, $\{\mathbf{v}_{j}\}_{j=1}^{k}$
  \end{algorithmic}
  \caption{Simple Prototypal Regression - Fitting}\label{algorithm:spr_fit}
\end{algorithm}

\begin{algorithm}[t]
  \begin{algorithmic}[1]
    \Require Value $\mathbf{x}_0$ of the predictor, prototypes $\{\mathbf{u}_{j}\}_{j=1}^{k}$ and $\{\mathbf{v}_{j}\}_{j=1}^{k}$ for predictor and response respectively, penalty coefficient $\lambda$.
    \Ensure Predicted $\hat{\mathbf{y}}_0$.
    \State $\displaystyle (a_{j}) \gets \argmin_{\substack{a_{j} \geq 0\\a_{1} + \cdots + a_{k}=1}}\norm{\mathbf{x}_0 - \sum_{j=1}^{k} a_{j} \mathbf{u}_j}^2 + \lambda \sum_{j=1}^{k}a_{j}\norm{\mathbf{x}_{0}- \mathbf{u}_{j}}^2$
    \State $\hat{\mathbf{y}}_{0} \gets a_{1} \mathbf{v}_1 + \cdots + a_{n} \mathbf{v}_{n}$
    \State \Return $\hat{\mathbf{y}}_{0}$
  \end{algorithmic}
  \caption{Simple Prototypal Regression - Prediction}\label{algorithm:spr_predict}
\end{algorithm}

\subsection{Multiple Prototypal Regression}

Multiple prototypal regression estimates the response $\mathbf{y}$ using $m$ predictors $\{\mathbf{x}^{(l)}\}_{l=1}^{m}$ (again, both the response and each of the predictors can be vectorial.) As in simple prototypal regression, it finds prototypes for $\mathbf{x}^{(l)}$ and $\mathbf{y}$ and builds the regression function on prototypes.

The prototypes of $\mathbf{x}^{(l)}$ still come from direct prototypal analysis, i.e. solving Equation \ref{eqn:pa} for each $\{\mathbf{x}^{(l)}_i\}_{i=1}^{n}$. Each predictor has $k_l$ prototypes and penalty coefficient $\lambda_l$, these need not be the same across predictors. When finding prototypes for $\mathbf{y}$, we weight the prototypes of each $\mathbf{x}^{(l)}$ by an importance coefficient. Denoting by $\mathbf{u}_{j}^{(l)}$ the prototypes of $\mathbf{x}^{(l)}$, by $\mathbf{v}_{j}^{(l)}$ the prototypes of $\mathbf{y}$ and by $\tau_l$ the importance coefficient corresponding to the $l$-th predictor, the regression function $\hat{\mathbf{f}}$ in multiple prototypal regression is given by
\begin{equation}
  \label{eqn:mpr:predict}
  \hat{\mathbf{f}}(\mathbf{x}_0) = \sum_{l=1}^{m} \tau_l \sum_{j=1}^{k_l} a_{j0}^{(l)} \mathbf{v}_j^{(l)},
\end{equation}
where $a_{j0}^{(l)}$ are the barycentric coordinates of $\mathbf{x}_0^{(l)}$ in prototypal analysis as in Equation \ref{eqn:spr:reconstruct}.

The importance coefficients $\tau_l$ in Equation \ref{eqn:mpr:predict} are non-negative and add up to one. Both the importance coefficients and the prototypes of $\mathbf{y}$ are obtained by minimizing the squared errors of Equation \ref{eqn:mpr:predict} on the data: denoting by $a_{ji}^{(l)}$ the weight of $\mathbf{u}_{j}^{(l)}$ for reconstructing $\mathbf{x}_{i}^{(l)}$,
\begin{equation}
  \label{eqn:mpr:prototype_y}
 \mathbf{v}_j^{(l)} = \sum_{i=1}^{n} c_{ij}^{(l)} \mathbf{y}_i, \quad c, \tau = \argmin_{\substack{c_{hj}^{(l)}, \tau_l \geq 0\\c_{1j}^{(l)} + \cdots + c_{nj}^{(l)}=1\\\tau_1 + \cdots + \tau_m=1}}\sum_{i=1}^{n}{\norm{\mathbf{y}_i - \sum_{l=1}^{m} \tau_l \sum_{j=1}^{k_l} a_{ji}^{(l)} \sum_{h=1}^{n} c_{hj}^{(l)} \mathbf{y}_h}^2}.
\end{equation}
Here the optimization is carried out through the alternate minimization over the $c$ and $\tau$.

\begin{algorithm}[t]
  \begin{algorithmic}[1]
    \Require Predictor data $\{\mathbf{x}^{(1)}_{i}\}_{i=1}^{n}, \cdots, \{\mathbf{x}^{(m)}_{i}\}_{i=1}^{n}$, response data $\{\mathbf{y}_{i}\}_{i=1}^{n}$, number of prototypes $k_1, \cdots, k_m$, penalty coefficient $\lambda_1, \cdots, \lambda_m$.
    \Ensure Prototypes $\{\mathbf{u}_{j}^{(1)}\}_{j=1}^{k_1}, \cdots, \{\mathbf{u}_{j}^{(m)}\}_{j=1}^{k_m}$ for predictors and $\{\mathbf{v}_{j}^{(1)}\}_{j=1}^{k_1}, \cdots, \{\mathbf{v}_{j}^{(m)}\}_{j=1}^{k_m}$ for response, importance coefficients $\tau_1,\cdots,\tau_m$.
    \For{$l=1,\cdots,m$}
    \State $\displaystyle (a_{ji}^{(l)}) , (b_{hj}^{(l)}) \gets$ \par $\displaystyle \argmin_{\substack{a_{ji}^{(l)} \geq 0, b_{lj}^{(l)} \geq 0\\a_{1i}^{(l)} + \cdots + a_{ki}^{(l)}=1\\b_{1j}^{(l)} + \cdots + b_{nj}^{(l)}=1}}{\sum_{i=1}^{n}{\norm{\mathbf{x}_i^{(l)} - \sum_{j=1}^{k_l} a_{ji}^{(l)} \sum_{h=1}^{n} b_{hj}^{(l)} \mathbf{x}_h^{(l)}}}^2 + \lambda_l \sum_{i=1}^{n}\sum_{j=1}^{k}a_{ji}^{(l)}\norm{\mathbf{x}_{i}^{(l)}-\sum_{h=1}^{n} b_{hj}^{(l)} \mathbf{x}_h^{(l)}}^2}$
    \For{$j=1,\cdots,k_l$}
    \State $\mathbf{u}_{j}^{(l)} \gets b_{1j}^{(l)} \mathbf{x}_1^{(l)} + \cdots + b_{nj}^{(l)} \mathbf{x}_{n}^{(l)}$
    \EndFor
    \EndFor
    \State $\displaystyle (c_{hj}^{(l)}), (\tau_l) \gets \argmin_{\substack{b_{hj}^{(l)}, \tau_l \geq 0\\c_{1j}^{(l)} + \cdots + c_{nj}^{(l)}=1\\\tau_1 + \cdots + \tau_m=1}}\sum_{i=1}^{n}{\norm{\mathbf{y}_i - \sum_{l=1}^{m} \tau_l \sum_{j=1}^{k_l} a_{ji}^{(l)} \sum_{h=1}^{n} c_{hj}^{(l)} \mathbf{y}_h}^2}$
    \For{$l=1,\cdots,m$}
    \For{$j=1,\cdots,k_l$}
    \State $\mathbf{v}_{j}^{(l)} \gets c_{1j}^{(l)} \mathbf{y}_1 + \cdots + c_{nj}^{(l)} \mathbf{y}_{n}$
    \EndFor
    \EndFor
    \State \Return $\{\mathbf{u}_{j}^{(1)}\}_{j=1}^{k_1}, \cdots, \{\mathbf{u}_{j}^{(m)}\}_{j=1}^{k_m}, \{\mathbf{v}_{j}^{(1)}\}_{j=1}^{k_1}, \cdots, \{\mathbf{v}_{j}^{(m)}\}_{j=1}^{k_m},\tau_1,\cdots,\tau_m$
  \end{algorithmic}
  \caption{Multiple Prototypal Regression - Fitting}\label{algorithm:mpr_fit}
\end{algorithm}

\begin{algorithm}[t]
  \begin{algorithmic}[1]
    \Require Values $\mathbf{x}_0=\left(\mathbf{x}_0^{(1)}, \cdots, \mathbf{x}_0^{(m)}\right)$ of the predictors, prototypes $\{\mathbf{u}_{j}^{(1)}\}_{j=1}^{k_1}, \cdots, \{\mathbf{u}_{j}^{(m)}\}_{j=1}^{k_m}$ for predictors and $\{\mathbf{v}_{j}^{(1)}\}_{j=1}^{k_1}, \cdots, \{\mathbf{v}_{j}^{(m)}\}_{j=1}^{k_m}$ for response, importance coefficients $\tau_1,\cdots,\tau_m$, penalty coefficients $\lambda_1, \cdots, \lambda_m$.
    \Ensure Predicted $\hat{\mathbf{y}}_0$.
    \For{$l=1,\cdots,m$}
    \State $\displaystyle (a_{j}^{(l)}) \gets \argmin_{\substack{a_{j}^{(l)} \geq 0\\a_{1}^{(l)} + \cdots + a_{k}^{(l)}=1}}\norm{\mathbf{x}_0^{(l)} - \sum_{j=1}^{k_l} a_{j}^{(l)} \mathbf{u}_j^{(l)}}^2 + \lambda_l \sum_{j=1}^{k_l}a_{j}^{(l)}\norm{\mathbf{x}_{0}^{(l)}- \mathbf{u}_{j}^{(l)}}^2$
    \EndFor
    \State $\displaystyle \hat{\mathbf{y}}_{0} \gets \sum_{l=1}^{m} \tau_l \sum_{j=1}^{k_l} a_j^{(l)} \mathbf{v}_j^{(l)}$
    \State \Return $\hat{\mathbf{y}}_{0}$
  \end{algorithmic}
  \caption{Multiple Prototypal Regression - Prediction}\label{algorithm:mpr_predict}
\end{algorithm}

\subsection{Applications}
\subsubsection{Iris Flowers}
We apply multiple prototypal regression to the data set for classification of Iris into species introduced by \citet{fisher1936use}. This includes three Iris species with four features for each flower: sepal length, sepal width, petal length and petal width.
In this example, we treat the sepal and petal dimensions as two two-dimensional predictors and one-hot encode the three species as $(1,0,0)$, $(0,1,0)$ and $(0,0,1)$. Multiple prototypal regression predicts a probability vector given the sepal and petal features. The species with highest probability is then adopted as predicted label.

There are 150 samples in the Iris data set with 50 samples for each species. Using stratified sampling, we randomly split the samples into a training set of 105 samples and a test set of 45 samples. By grid search with cross validation on the training data, we pick the number of prototypes to be $11$ and the penalty coefficient to be $0.1$ for both features. The accuracy scores on the training and testing sets are shown in Table \ref{tab:iris_score}.

\begin{figure}[ptb]
  \begin{center}
    \input{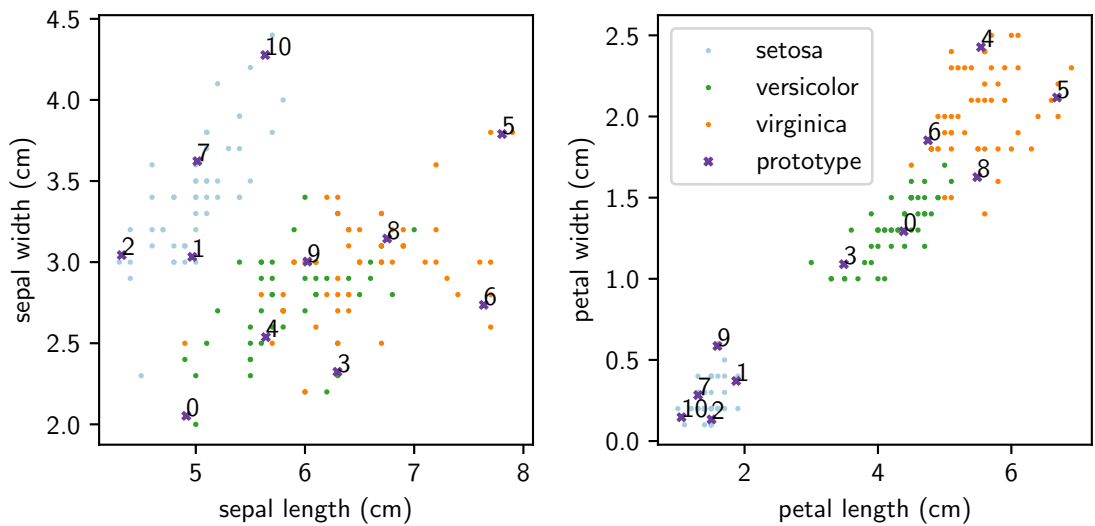}
  \end{center}
  \caption{Sepal dimensions and petal dimensions of Iris flowers and their prototypes.}
  \label{fig:iris_predictor}
\end{figure}

\begin{figure}[ptb]
  \begin{center}
    \input{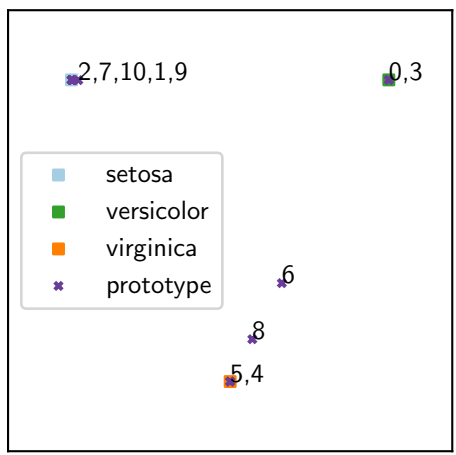}
  \end{center}
  \caption{Species of Iris flowers and prototypes corresponding to petal dimensions. This plot of the three-dimensional object $(P_1, P_2, P_3)$ is represented here in barycentric coordinates, where the three vertices of the triangle correspond to the three species.}
  \label{fig:iris_response}
\end{figure}

The Iris data set and the prototypes of the sepal and petal dimensions are shown in Figure \ref{fig:iris_predictor}. Figure \ref{fig:iris_predictor} suggests the petal dimensions are more informative than the sepal's for the classification task. This agrees with the importance coefficients of prototypal regression, which are $3\times10^{-7}$ and $0.9999997$ for the sepal and petal dimensions respectively. Figure \ref{fig:iris_response} shows the responses of this classification problem and the prototypes of the responses corresponding to the petal dimensions.

\begin{table}[t]
  \begin{center}
    \begin{tabular}{ | c | c | c | }
      \hline      
      & training score & test score \\
      \hline
      prototypal regression & 0.96 & 1.00\\
      \hline
    \end{tabular}
  \end{center}
  \caption{Accuracy score on Iris flowers data set.}
  \label{tab:iris_score}
\end{table} 

\section{Kernels and Extension to Probability Distributions}
\label{sec:kernel}
\subsection{Prototypal Learning with Kernels}
\label{sec:kernel_pl}
Archetypal analysis, prototypal analysis and prototypal regression involve the data only through the pairwise inner products 
\begin{equation*}
  \langle \mathbf{x}_i, \mathbf{x}_j \rangle \ \text{and} \ \langle \mathbf{y}_i, \mathbf{y}_j \rangle
\end{equation*}
as follows from expanding the squared norms in Equation \ref{eqn:aa}, \ref{eqn:pa}, \ref{eqn:spr:prototype_y} and \ref{eqn:mpr:prototype_y}.
Hence we can extend all three to reproducing kernel Hilbert spaces. Choosing a symmetric and positive semidefinite kernel function $K$, the map from $\mathbf{x}_i$ to $h(\mathbf{x}_i)=K(\cdot,\mathbf{x}_i)$ yields the inner product
\begin{equation*}
  \label{eqn:rkhs:inner}
  \langle h(\mathbf{x}_i),h(\mathbf{x}_j) \rangle = K(\mathbf{x}_i,\mathbf{x}_j),
\end{equation*}
which replaces the inner products in Equation \ref{eqn:aa}, \ref{eqn:pa}, \ref{eqn:spr:prototype_y} and \ref{eqn:mpr:prototype_y}, and extends archetypal analysis, prototypal analysis and prototypal regression to a (potentially infinite-dimensional) reproducing kernel Hilbert space.


\subsection{Prototypal Learning on Distributions through Kernel Embedding}
Probability distributions or samples thereof  can also be mapped to a reproducing kernel Hilbert space via kernel embedding \citep[see][]{berlinet2011reproducing,gretton2006kernel,smola2007hilbert,sriperumbudur2010hilbert,SejdinovicGSFJ2012,muandet2017kernel}. With a symmetric, positive semidefinite kernel function $K(\cdot,\cdot)$ on $\mathcal{X}\times\mathcal{X}$, the kernel embedding $g$ maps a probability measure $\mu(\cdot)$ on $\mathcal{X}$ to a reproducing kernel Hilbert space through
\begin{equation}
  \label{eqn:kernel_embedding}
  \mu(\cdot) \mapsto g(\mu(\cdot)) = \int_{\mathcal{X}}{K(\cdot,x) d\mu(x)},
\end{equation}
with induced inner product given by
\begin{equation}
  \label{eqn:kernel_embedding_inner_product}
  \langle g(\mu_1(\cdot)), g(\mu_2(\cdot)) \rangle = \int_{\mathcal{X}\times\mathcal{X}}{K(x_1,x_2) d\mu_1(x_1) d\mu_2(x_2)}.
\end{equation}

Kernel embedding does not necessarily yield an injective map; \citet{sriperumbudur2010hilbert} give several criteria for whether a kernel induces an injective embedding for distributions on $\mathbb{R}^{d}$ and $\mathbb{T}^d$. Some commonly used kernels on $\mathbb{R}^d$ for injective kernel embeddings are listed in Table \ref{tab:kernels}. The Gaussian, Laplacian and $B_{2n+1}$-spline kernels are shown to induce injective embeddings in \citet{sriperumbudur2010hilbert}. The energy distance kernel induces an embedding well-defined on distributions with finite first moment. The energy distance $D_{\text{ED}}$ \citep{szekely2013energy,rizzo2016energy}:
\begin{equation*}
  \label{eqn:energy_distance}
  \begin{split}
    D_{\text{ED}}^2(\mu_1(\cdot),\mu_2(\cdot)) = &2\int_{\mathcal{X}\times\mathcal{X}}{\norm{x_1-x_2} d\mu_1(x_1) d\mu_2(x_2)} - \int_{\mathcal{X}\times\mathcal{X}}{\norm{x_1-x_2} d\mu_1(x_1) d\mu_1(x_2)}\\
    &- \int_{\mathcal{X}\times\mathcal{X}}{\norm{x_1-x_2} d\mu_2(x_1) d\mu_2(x_2)}
  \end{split}
\end{equation*}
is proved in \citet{klebanov2002class} to yield a metric, implying that the energy distance kernel induces an injective embedding.

Replacing the integrals in Equation \ref{eqn:kernel_embedding} and \ref{eqn:kernel_embedding_inner_product} by the corresponding empirical means gives the kernel embedding and induced inner product for samples of distributions. Given samples $\{\mathbf{x}_i\}_{i=1}^{n}$ of $\mu$, the kernel embedding for the empirical distribution $\hat{\mu}$ is
\begin{equation*}
  \label{eqn:empirical_kernel_embedding}
  \hat{\mu}(\cdot) \mapsto g(\hat{\mu}(\cdot)) = \frac{1}{n} \sum_{i=1}^{n}{K(\cdot,x_i)},
\end{equation*}
and given samples $\{\mathbf{x}_i^{(1)}\}_{i=1}^{n_1}$, $\{\mathbf{x}_i^{(2)}\}_{i=1}^{n_2}$ of $\mu_1$ and $\mu_2$, the induced inner product of the empirical distributions $\hat{\mu_1}$ and $\hat{\mu_2}$ is
\begin{equation*}
  \label{eqn:empirical_kernel_embedding_inner_product}
  \langle g(\hat{\mu}_1(\cdot)), g(\hat{\mu}_2(\cdot)) \rangle = \frac{1}{n_1 n_2} \sum_{i_1=1}^{n_1}{\sum_{i_2=1}^{n_2}{K(x_{i_1}^{(1)},x_{i_2}^{(2)})}}.
\end{equation*}
In general, the time complexity of evaluating the inner product is $O(n_1 n_2)$. For the Gaussian kernel, the time complexity for the inner product can be reduced to $O(n_1 + n_2)$ via the fast Gauss transform \citep{greengard1991fast} or the improved fast Gauss transform \citep{yang2003improved}. For the energy distance kernel on sorted samples of one-dimensional distributions, the time complexity of evaluating the inner product is $O(n_1 + n_2)$, as shown in Appendix \ref{app:energy distance}.

We can extend archetypal analysis, prototypal analysis and prototypal regression to distributions with the inner products induced by kernel embedding. In archetypal/prototypal analysis, the archetypes/prototypes are mixtures of the input distributions and their mixtures are used to reconstruct the input distributions. In prototypal regression, we can have distributions as predictors, responses or both. In multiple prototypal regression, we can blend numerical, categorical and distributional predictors.

\begin{table}[tb]
  \begin{center}
    \begin{tabular}{ | c c |}
      \hline
      kernel & $K(x,y)$\\
      \hline
      Gaussian & $e^{-\sigma\norm{x-y}^2}$\\
      Laplacian & $e^{-\sigma\norm{x-y}_1}$\\
      $B_{2n+1}$-spline & $\displaystyle\prod_{i=1}^{d}B_{2n+1}(x_i-y_i)$\\
      energy distance & $\norm{x} + \norm{y} - \norm{x-y}$\\
      \hline
    \end{tabular}
  \end{center}
  \caption{Some commonly used kernels on $\mathbb{R}^d$ for injective kernel embeddings. For $B_{2n+1}$-spline, $B_{2n+1}(x) = *_1^{(2n+2)}\mathbf{1}_{[-\frac{1}{2},\frac{1}{2}]}(x)$, where the symbol $*_1^{(2n+2)}$ represents the $(2n + 2)$-fold convolution.}
  \label{tab:kernels}
\end{table} 

\subsection{Applications}
\subsubsection{Smartphone-based Human Activities Recognition Data Set}
The smartphone-based human activities recognition data set from \citet{anguita2013public} and \citet{reyes2016transition} contains activity data collected by smartphone's inertial sensors. In their experiments, 30 volunteers conducted 6 activities: walking, walking upstairs, walking downstairs, sitting, standing and laying while wearing a wrist-mounted smartphone.
The data set contains raw and processed data. The raw data are the triaxial signals from the accelerometer and the gyroscope of smartphones at a constant rate of 50Hz for each activity. The processed data include statistics, such as the mean, standard deviation and auto correlation of the raw signals, and other data, such as the magnitude and the fast Fourier transform of the raw signals.

\citet{anguita2013public} and \citet{reyes2016transition} use the processed data to classify the activities. We use the raw data instead, i.e. the triaxial signals from the accelerometer and gyroscope. Each trial in the raw data set contains two three-dimensional time series of the accelerometer and the gyroscope respectively and a label of the activity. We divide the data set into a training data set of 772 trials and a test data set of 84 trials. Multiple prototypal regression is applied for this classification task. The samples of triaxial signals from the accelerometer and the gyroscope are the two predictors in multiple prototypal regression and energy distance kernel is used for kernel embedding. The labels are binarized via one-hot encoding. The number of prototypes is set to be $70$ and the penalty coefficient is set to be $1$ for both predictors.
We achieve a $97.62\%$ accuracy on the testing subset. The confusion matrix for the test data is shown in Table \ref{tab:activity_confusion}, the importance coefficients are listed in Table \ref{tab:activity}.

\begin{table}[tb]
  \begin{center}
    \begin{tabular}{ | c | c | c | c | c | c | c |}
      \hline
      & walk & upstairs & downstairs & sit & stand & lay\\
      \hline
      walk       & 12 & 0  & 0  & 0  & 0  & 0\\
      \hline
      upstairs   & 1  & 17 & 0  & 0  & 0  & 0\\
      \hline
      downstairs & 0  & 0  & 18 & 0  & 0  & 0\\
      \hline
      sit        & 0  & 0  & 0  & 11 & 1  & 0\\
      \hline
      stand      & 0  & 0  & 0  & 0  & 12 & 0\\
      \hline
      lay        & 0  & 0  & 0  & 0  & 0  & 12\\
      \hline
    \end{tabular}
  \end{center}
  \caption{Confusion matrix of multiply prototypal regression on smartphone-based human activities recognition data set. The rows are the actual classes and the columns are the predicted classes.}
  \label{tab:activity_confusion}
\end{table} 

\begin{table}[tb]
  \begin{center}
    \begin{tabular}{ | c | c | c | c | c | c | c |}
      \hline
      &accelerometer &gyroscope\\
      \hline
      importance coefficients & 0.44 & 0.56\\
      \hline
    \end{tabular}
  \end{center}
  \caption{Importance coefficients of multiply prototypal regression on smartphone-based human activities recognition data set.}
  \label{tab:activity}
\end{table} 

\subsubsection{EPA Outdoor Air Quality Data Set}
The EPA Outdoor Air Quality Data \citep{epa2017dataset} collects pollutant and meteorological data at outdoor monitors across the United States, Puerto Rico, and the U. S. Virgin Islands. This data set contains hourly data of criteria gases (Ozone, \ce{SO2}, \ce{CO} and \ce{NO2}), toxics and precursors (HAPs, VOCs, \ce{NONOxNOy} and lead), particulates (PM2.5 FRM/FEM Mass, PM2.5 non FRM/FEM Mass, PM10 Mass and PM2.5 Speciation) and meteorological data (winds, temperature, barometric pressure, relative humidity and dew point).

\begin{figure}[tb]
  \begin{center}
    \input{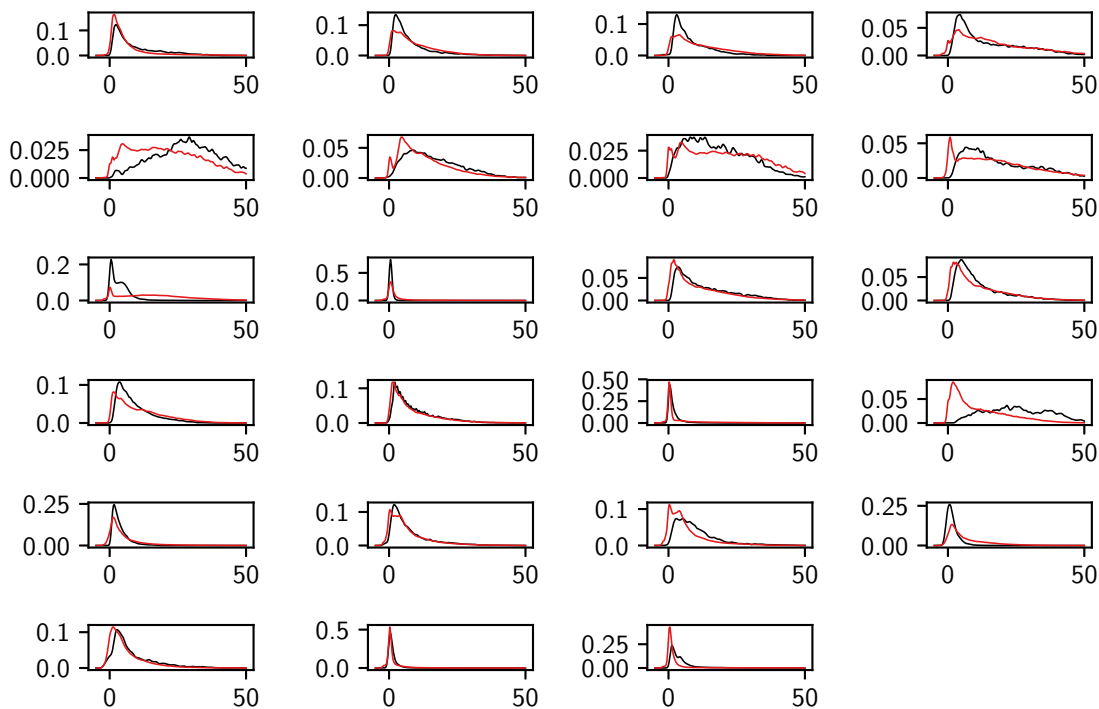}
  \end{center}
  \caption{Out-of-sample prediction of \ce{NO2} density distribution. The black curves are the true \ce{NO2} distributions at each station and the red curves are the predicted \ce{NO2} distributions by multiple prototypal regression.}
  \label{fig:epa_airquality}
\end{figure}

We use multiple prototypal regression to estimate the distributions of the nitrogen dioxide (\ce{NO2}) density from the geophysical locations (the latitude and longitude of the stations) and the distributions of the meteorological data. The meteorological data that we use are the one-dimensional distribution of wind speed, the one-dimensional distribution of wind direction and one-dimensional distribution of outdoor temperature. The training data set contains the data collected in the year 2016 at $200$ stations and the test data set contains the data collected in the same year at $23$ other stations. We use the energy distance kernel for embedding. The number of prototypes is set to $40$ and the penalty coefficient to $0.1$ for all predictors. The importance coefficients are listed in Table \ref{tab:epa_airquality} and the out-of-sample predictions are illustrated in Figure \ref{fig:epa_airquality}.

\begin{table}[tb]
  \begin{center}
    \begin{tabular}{ | c | c | c | c | c | c | c |}
      \hline
      &location &temperature &wind direction &wind speed\\
      \hline
      importance coefficients & 0.23 & 0.40 & 0.13 & 0.24\\
      \hline
    \end{tabular}
  \end{center}
  \caption{Importance coefficients of multiply prototypal regression on EPA outdoor air quality data set.}
  \label{tab:epa_airquality}
\end{table} 

\section{Conclusions}
\label{sec:conclusion}
We have proposed and developed prototypal analysis and regression, two robust extensions of archetypal analysis. In addition, we have shown how these methodologies can be extended via kernel embedding to handle learning problems where the data points are probability distributions known through samples. Here the interpretability associated with the convex combinations involved is clearest, as these combinations can be interpreted as mixtures of distributions.

Prototypal analysis adds to the objective function of archetypal analysis a term that penalizes the use of distant prototypes for the reconstruction of data points. It can be regarded of as an interpolation between archetypal analysis---corresponding to a zero value of the penalization parameter $\lambda$---and k-means, which arises as $\lambda \rightarrow \infty$. This adds robustness to outliers and a sense of locality, which becomes particularly useful when the methodology is used for regression.

We illustrate through real-life examples the applicability of the procedure, particularly to scenarios that blend numerical and distributional features or that have probability distributions as labels to predict.

\acks{The work of E. G. Tabak was partially supported by grants from the Office of Naval Research and the Mathematical Division of the National Science Foundation.}

\appendix
\section{Energy Distance Kernel of One-Dimensional Distributions}
\label{app:energy distance}
The energy distance kernel on distributions $\mu,\nu$ can be estimated using their samples $\{x_{i}\}_{i=1}^{n_x},\allowbreak\{y_{j}\}_{i=1}^{n_y}$ through the empirical mean:
\begin{equation}
\label{eqn:energy_estimate}
k_{\text{ED}}(\mu,\nu) \approx \frac{1}{n_{x}} \sum_{i=1}^{n_{x}} \norm{x_{i}} + \frac{1}{n_{y}} \sum_{j=1}^{n_{y}} \norm{y_{j}} -\frac{1}{n_{x}n_{y}} \sum_{i=1}^{n_{x}} \sum_{j=1}^{n_{y}}  \norm{x_{i} - y_{j}}.
\end{equation}
The time complexity of evaluating Equation \ref{eqn:energy_estimate} is $O(n_x n_y)$.

For one-dimensional distributions, the time complexity of evaluating Equation \ref{eqn:energy_estimate} can be reduced to the linear $O(n_x + n_y)$ when the samples $\{x_{i}\}_{i=1}^{n_x},\{y_{j}\}_{i=1}^{n_y}$ are sorted, as illustrated in Algorithm \ref{algorithm:energy}.
The intuition behind is that each term in $\sum_{i=1}^{n_{x}} \sum_{j=1}^{n_{y}}  \| x_{i} - y_{j} \|$ can be expanded into
\begin{equation*}
\| x_{i} - y_{j} \| = \mathbf{1}_{x_i > y_j} (x_i-y_j) - \mathbf{1}_{x_i \leq y_j} (x_i-y_j) = (\mathbf{1}_{x_i > y_j} - \mathbf{1}_{x_i \leq y_j}) x_i + (\mathbf{1}_{x_i \leq y_j} - \mathbf{1}_{x_i > y_j}) y_j,
\end{equation*}
yielding
\begin{equation}
\label{eqn:energy_estimate_linear}
\sum_{i=1}^{n_{x}} \sum_{j=1}^{n_{y}}  \| x_{i} - y_{j} \| = \sum_{i=1}^{n_{x}} \left[ \sum_{j=1}^{n_{y}} (\mathbf{1}_{x_i > y_j} - \mathbf{1}_{x_i \leq y_j}) \right] x_i  +  \sum_{j=1}^{n_{y}} \left[ \sum_{i=1}^{n_{x}} (\mathbf{1}_{x_i \leq y_j} - \mathbf{1}_{x_i > y_j}) \right] y_j.
\end{equation}
Equation \ref{eqn:energy_estimate_linear} implies that we only need to count how many $y_j$'s are smaller than each $x_i$ and how many $x_i$'s are smaller than each $y_j$. If the samples are sorted, this counting can be done in linear time.

\begin{algorithm}[t]
  \begin{algorithmic}[1]
    \Require Sorted samples $\{x_{i}\}_{i=1}^{n_x}, \{y_{j}\}_{j=1}^{n_y}$ of 1D distributions $\mu,\nu$.
    \Ensure Empirical estimation of energy distance kernel $k_{\text{ED}}(\mu,\nu)$
    \State $\text{sum}_{x} \gets 0, \text{sum}_{y} \gets 0, i \gets 1, j \gets 1$\;
    \While{$i \leq n_{x}$ \textbf{and}  $j \leq n_{y}$} \If{$x_{i} \leq y_{j}$}
    \State $\text{sum}_{x} \gets \text{sum}_{x} + \{ (j-1) - \left[n_{y}-(j-1)\right] \} x_{i}$
    \State $i \gets i + 1$
    \Else
    \State $\text{sum}_{y} \gets \text{sum}_{y} + \{ (i-1) - \left[n_{x}-(i-1)\right] \} y_{j}$
    \State $j \gets j + 1$
    \EndIf
    \EndWhile
    \If{$i>n_{x}$}
    \State $\text{sum}_{y} \gets \text{sum}_{y}+n_x\sum\limits_{k=j}^{n_{y}}y_{k}$
    \Else
    \State $\text{sum}_{x} \gets \text{sum}_{x}+n_y\sum\limits_{k=i}^{n_{x}}x_{k}$
    \EndIf
    \State $k_{\text{ED}}(\mu,\nu) \gets \left(\sum\limits_{k=1}^{n_{x}}x_{k}\right)/n_x + \left(\sum\limits_{k=1}^{n_{y}}y_{k}\right)/n_y - (\text{sum}_{x} + \text{sum}_{y})/(n_x n_y)$
    \State \Return $k_{\text{ED}}(\mu,\nu)$
  \end{algorithmic}
  \caption{Energy Distance Kernel of 1D distributions}\label{algorithm:energy}
\end{algorithm}

\vskip 0.2in
\bibliography{collection}

\end{document}